\documentclass[10pt,twocolumn,letterpaper]{article}

\usepackage[pagenumbers]{wacv} 

\usepackage{graphicx}
\usepackage{amsmath}
\usepackage{amssymb}
\usepackage{booktabs}

\newcommand{\CR}[1]{{\color{black}{{#1}}}}

\usepackage[pagebackref,breaklinks,colorlinks]{hyperref}

\usepackage[capitalize]{cleveref}
\crefname{section}{Sec.}{Secs.}
\Crefname{section}{Section}{Sections}
\Crefname{table}{Table}{Tables}
\crefname{table}{Tab.}{Tabs.}

\def\wacvPaperID{861} 
\def\confName{WACV}
\def\confYear{2025}

\usepackage{fancyhdr}
\fancypagestyle{firstpage}{
  \fancyhf{} 
  \fancyhead[C]{To appear in proceedings of the 2025 IEEE/CVF Winter Conference on Applications of Computer Vision (WACV) \\ The final publication will be available soon.} 
  \headsep = 0.7cm
  \fancyfoot[C]{1}
}

\begin{document}

\title{Dense Scene Reconstruction from Light-Field Images\\ Affected by Rolling Shutter}

\author{
Hermes McGriff$^{~1,3}$ \hspace{0.2in}
Renato Martins$^{1,2}$   \hspace{0.2in}
Nicolas Andreff$^{~3}$   \hspace{0.2in}
Cédric Demonceaux$^{1,2}$ \vspace{0.05in}\\
$^1$Université de Bourgogne, CNRS UMR 6303 ICB \hspace{0.04in}
$^2$Université de Lorraine, CNRS, Inria, LORIA \hspace{0.04in}\\
$^3$Université de Franche-Comté, CNRS UMR 6174 FEMTO-ST\\
{\tt\footnotesize \{hermes.mc-griff,renato.martins,cedric.demonceaux\}@u-bourgogne.fr, nicolas.andreff@univ-fcomte.fr}
}

\maketitle

\thispagestyle{firstpage}
\begin{abstract}

This paper presents a dense depth estimation approach from light-field (LF) images that is able to compensate for strong rolling shutter (RS) effects. Our method estimates RS compensated views and dense RS compensated disparity maps. We present a two-stage method based on a 2D Gaussians Splatting that allows for a ``render and compare" strategy with a point cloud formulation. In the first stage, a subset of sub-aperture images is used to estimate an RS agnostic 3D shape that is related to the scene target shape ``up to a motion". In the second stage, the deformation of the 3D shape is computed by estimating an admissible camera motion. We demonstrate the effectiveness and advantages of this approach through several experiments conducted for different scenes and types of motions. Due to lack of suitable datasets for evaluation, we also present a new carefully designed synthetic dataset of RS LF images. The source code, trained models and dataset will be made publicly available at: \url{https://github.com/ICB-Vision-AI/DenseRSLF}.

\end{abstract}

\section{Introduction}
\begin{figure}[t]
\centering
\includegraphics[width=\linewidth]{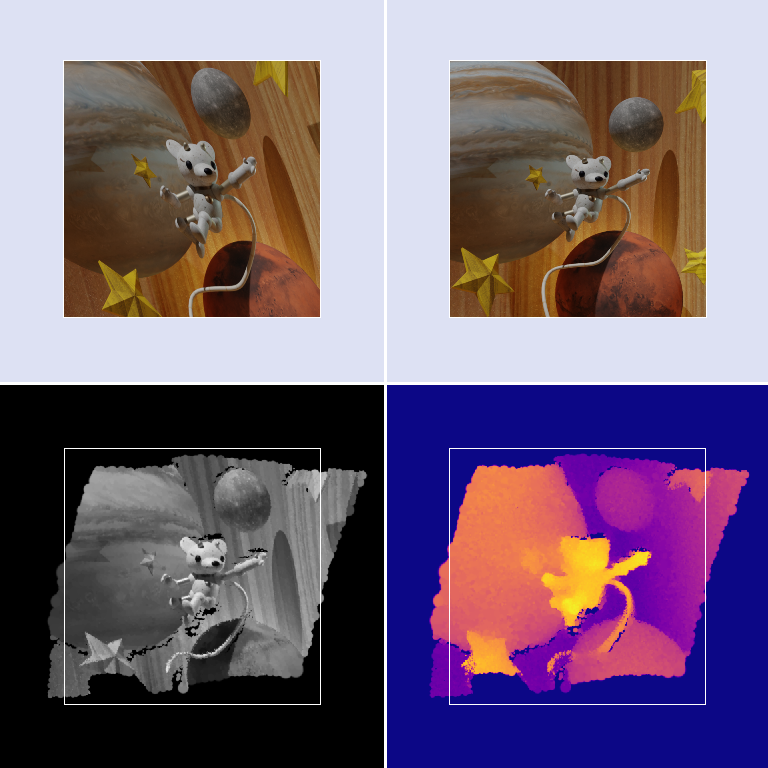}
\caption{\textbf{Dense depth and motion compensation example from a light-field image affected by RS distortions.}
The RS deformed central sub-aperture view (used as input) is shown in the \textit{top left}. The \textit{top right} view shows the corresponding view with no RS effect. Our method results are shown in the second row, with the rendered compensated appearance model (\textit{bottom left}) and disparity (\textit{bottom right}) generated from the set of learned 2D Gaussians.}\vspace*{-0.5cm}
\label{intro}
\end{figure}
Light-field cameras (\emph{a.k.a.} plenoptic cameras) are capable, thanks to their optical system, of recovering a 4D light field (LF) from a single capture. LFs have gained a strong popularity in the vision and graphics community notably since the apparition of neural radiance fields (NeRFs, e.g., ~\cite{mildenhall2021nerf,yu2021pixelnerf,barron2021mip}) and more recently 3D Gaussian Splatting~\cite{kerbl20233d}. In the LF literature, images are generally considered to be acquired from a global shutter sensor, \ie, under the assumption that all pixels are measured at the same time instant. But most consumer available cameras are equipped with rolling shutter (RS) sensors.
\begin{figure*}[t!]
\centering
\includegraphics[width=0.9\linewidth]{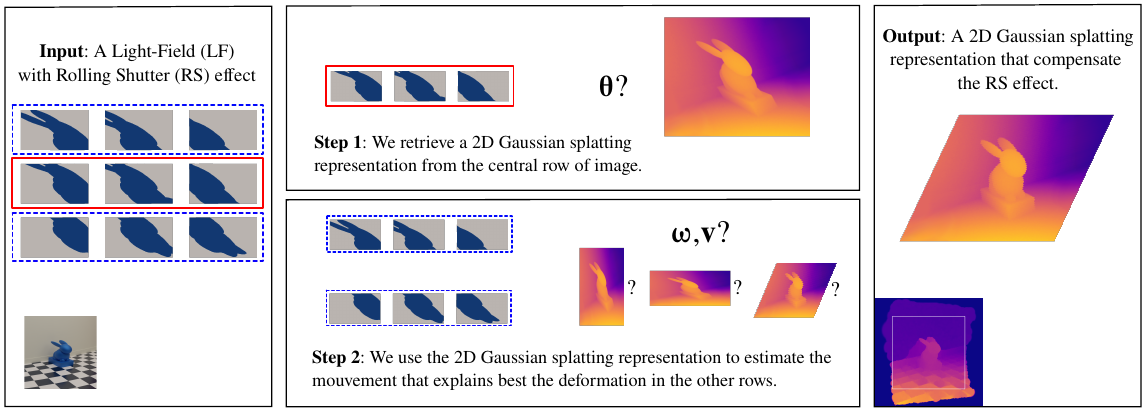}
\caption{\textbf{Overview of the main components of our two-stage method}. In the first stage, we estimate a 2D Gaussian Splatting representation from a sub-set of SAIs. In the second stage, using the ability to render and compare thanks to the 2D Gaussian Splatting representation, we estimate the motion parameters in order to recover an undistorted representation of the scene.}
\label{fig-method}
\end{figure*}
The particularity of RS cameras is that the acquisition of pixels is sequential (line by line) and, in the presence of either a camera motion or dynamic objects during the acquisition, the obtained image appears deformed by what is called the RS effect. It prevents the disambiguation of the impact of motion and shape in the RS imaging process. Ait-Aider~\cite{ait2006simultaneous} showed that we could leverage these deformations for motion estimation if both object shape and 2D-3D correspondences are known. More recently, we have shown in ~\cite{HermesWACV24} that in the case of an RS light field, it is possible to estimate jointly the shape and the motion without any prior knowledge of the scene. However, this approach still requires 2D point correspondences between subaperture views of the LF. This constraint limits the density of object points and generates sparse reconstructions of distinctive object regions such as corners. In this paper, we propose a new method that is capable of producing dense reconstructions without the need of priors neither in the shape of the scene nor in the velocity profile. Different than most RS correcting techniques for monocular setups, we only require one capture to estimate both depth and motion and do not rely on multiple acquisitions~\cite{liu2020deep, lao2021solving, karpenko2011digital}. Our method explores the potential offered by a Gaussian Splatting scene representation adapted to LF images to avoid performing correspondences. The designed approach presents different changes compared to existing Gaussian Splatting techniques~\cite{kerbl3Dgaussians} such as the use of 2D Gaussians, which are motivated by the specificity of the Light Fields acquired with a plenoptic camera. 
The method presented in this paper is a two-stage technique to obtain a dense representation of the scene, in terms of depth and appearance, jointly with motion parameters in order to compensate for the RS effect. These motion parameters are able to model the 3D deformation from the ``3D RS effect" and the motion of the camera. Our method outputs dense disparity maps and a compensated RS corrected light field as shown in \cref{intro}. Different experiments show the ability of this method to obtain dense 3D representations of the scene, even in the case of important RS effect distortions. 
The main technical contributions of this paper are as follows:

\begin{itemize}
    \item We present a new 2D Gaussian Splatting representation carefully designed to leverage geometric properties of LF images. This dense appearance and geometric representation is exploited in a differentiable rendering of the observed scene.
    \item An optimization-based approach is then designed to account for eventual rolling-shutter effects. We estimate motion parameters to correct the geometry, with a simple modelling of the deformation in the 3D space induced by RS in the light-field.
    \item Due to the lack of publicly existing LF datasets affected by rolling shutter, we present a new light-field dataset with rolling shutter effects. This dataset is designed to allow the evaluation on scenes with highly textured surfaces, while containing different levels of geometric details. 
\end{itemize}
We will publicly release the source code, trained models, dataset and test samples following the review process.

\section{Related work}

Leveraging the capabilities of rolling-shutter and light-field cameras for scene analysis has generated growing interest, although most current methods tend to utilize each modality independently. Ait-Aider \etal~\cite{ait2006simultaneous} seminal work was the first to  demonstrate the potential of leveraging the RS effect to estimate object motion, but they assumed that both object 3D shape and 2D-3D correspondences where known in order to estimate the motion. Following works like Ait-Aider \etal~\cite{ait2009structure}, Saurer \etal~\cite{saurer2013rolling} and Lao \etal~\cite{lao2021solving} explored the duality of deformation and RS effects, particularly in the context of stereo vision. Recently, \CR{we}~\cite{HermesWACV24} have proposed an optimization-based method that estimates the motion and scene structure for LF cameras affected by RS. Although that approach does not require either a prior shape of the scene or 2D-3D correspondences, it still required 2D-2D image correspondences between sub-aperture images (SAI). Consequently, the obtained 3D models are sparse, in contrast to the proposed method of this paper. 

Recently, 3D Gaussian Splatting~\cite{kerbl3Dgaussians} has been introduced with promising results for efficient differentiable rendering. Various extensions and applications have been proposed beyond graphics rendering due to its properties~\cite{keetha2023splatam, splatCVPR2024} notably by allowing the increase of the observed scene or varying point densities (which were challenges with neural fields such as NeRFs). In this context, Seisakri \etal~\cite{seiskari2024gaussian} proposed a method to create a Gaussian-Splatting from rolling shutter affected images from a standard monocular camera. However, the motion estimation is made from visual-inertial odometry over multiple captures. In this paper, we propose an adapted 2D Gaussian splatting tailored to leverage the sensor characteristics of plenoptic cameras. We present a strategy that is capable of implicitly handling correspondences though the minimization of image intensities using this adapted Gaussian splatting representation. Therefore we are capable of not requiring any shape, motion or correspondence priors, and we can render RS compensated dense depth maps and appearance views.

\section{Method}
Our approach is composed of two main components to obtain dense scene representation estimates with compensated RS effect from a light field, as depicted in \cref{fig-method}. An introduction and background to light fields and plenoptic cameras is provided in the ``Supplementary Material". We start by estimating a dense representation of the scene with a 2D Gaussian splatting. This component provides location of 2D Gaussians in the central sub-aperture image, their disparity, their size, and their intensity value. We have performed several adaptations of the 3D Gaussian splatting in order to leverage the specificities of the LF images. In the second stage,  we use this representation in a multiview reprojection strategy to obtain the angular and linear velocities of the camera, through minimization of appearance intensity errors. This stage allows us to disambiguate the impact of motion and shape in the RS imaging process.

\subsection{Rolling shutter light field properties}
\begin{figure}[t]
\centering
\includegraphics[width=\linewidth]{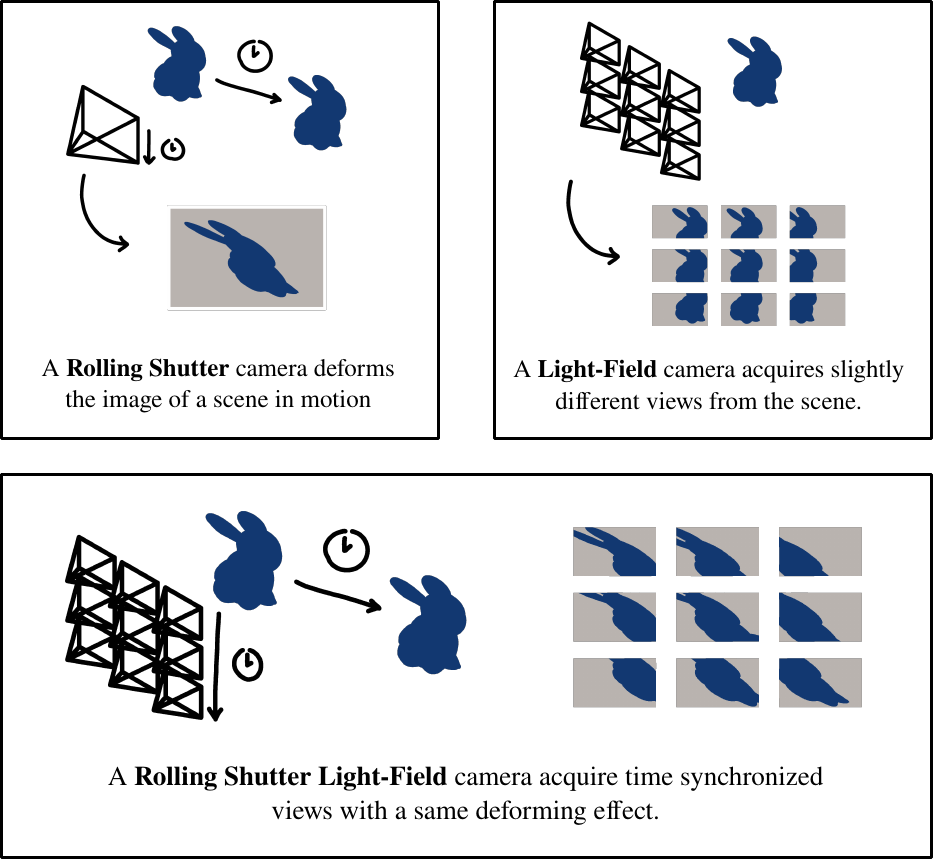}
\caption{\textbf{Properties of a rolling shutter light-field image (RSLF) with motion}. An RSLF sensor provides 4D data possessing the joint properties of an RS sensor (top left view) and of a Light-Field sensor (top right). It outputs RS affected sub-aperture images (SAI).\vspace{-0.5cm}}
\label{fig:sensor-setup}
\end{figure}
The proposed method takes advantage of the singular properties of the light fields acquired with a rolling-shutter device (as shown in \cref{fig:sensor-setup}). Firstly, each point of the scene is seen from different viewpoints given by the spatial array of virtual cameras of the light field (as shown in the upper right image of \cref{fig:sensor-setup}). It provides cues to recover the shape of the observed points. Secondly, a same point is measured at different time instants, providing cues to recover the motion (as shown in the top left view). However, these two information are entangled during acquisition. The main objective is then to design a method capable of leveraging these cues while disentangling shape and motion. 

There exist different designs of LF cameras \cite{wu2017light, zhou2021review}, but they all output 4D data. From this ``4D image'', one can extract a 2D array of 2D images, usually called Sub-Aperture Images (as shown in top right and bottom images of \cref{fig:sensor-setup}). An intrinsic property of the RS LF images is that the Sub-Aperture Images (SAI) are time synchronized 
(please see the ``Supplementary Material" for details).

\subsection{Disparity estimation and 3D RS effect} \label{sec:c_row}
Since we do not want to consider the motion of 3D points as disparity, we have to compute the disparity between points while they are seen at the same time instant. For that, we use only the SAI images of the central row (highlighted by a red box in left corner of \cref{fig-method}). By using only SAIs from the same row, we ensure that a given 3D point is projected in the same line in every SAI, which means at the same time instant.
With the disparity computed between this sub-set of SAIs, we have estimated a geometry affected by a ``3D rolling shutter effect''. Indeed, the estimated points are in the 3D space, but they are still observed at different times. But this deformed shape in correct ``up-to-a-motion''. Which means that it exists a deformation due to the motion that links the correct geometry and this 3D RS affected geometry. We will then use the rest of the light field to find this motion, and thus this deformation.
\subsection{2D Gaussian splatting representation}
We want to re-project the whole scene in order to compute a dense intensity re-projection error (without a matching of points) in a render-and-compare estimation scheme. We have decided to adapt 3D Gaussian Splatting~\cite{kerbl3Dgaussians} representation considering the properties of an LF camera setup. In LF camera images, the viewpoint of each SAI have small baselines and all information is ``aggregated" around the central view, conversely to typical operation conditions of NeRFs and 3D Gaussian Splatting. In this way the directions of the rays are concentrated in a narrow cone, which can lead to simplifications in the Gaussian splatting modelling. Notably, we can now assume diffuse Lambertian surfaces, and consider that the intensity values of the Gaussians are no longer dependent of the viewing direction. In the 3D Gaussian Splatting representation, the goal is to capture dense volumes of widely-sampled scenes. As aforementioned, the LFs captured with light field cameras (plenoptic cameras) sample the scene differently. This means that we are more motivated to represent surfaces than volumes. And we care less about view dependence of the Gaussian parameters for finding color and shape. 
We reduced the Gaussian formulation to its bare minimum, with 3 parameters for the center of a Gaussian $\mathbf{P}$ (equivalently to the 3D Gaussian Splatting), one parameter for the size of the Gaussian (instead of a $3\times3$ covariance matrix) and one for the intensity.\\ 
We use the disparity map estimated previously, in order to initialize a 2D Gaussian Splatting representation. Basically, we sample the central SAI at points that will be the 2D Gaussian centers (with more density in the regions of higher frequency) and by associating the intensity of the Gaussian with the pixel intensity at the Gaussian center, and disparity with the disparity value at this pixel in the disparity map. By rendering and alpha-blending (in inverse disparity order) of these Gaussians, we create a representation of the SAI. By displacing the center of the Gaussians with respect to their disparity, we can also render the scene from other view points (in a parallax logic). By rendering the scene at the view point of the other SAIs of the central line of SAIs, we minimize the difference between the real SAI and the rendered SAI by optimizing the coordinates of the center of the Gaussians, the intensity of the Gaussians, and the size of the Gaussians. After this fine tuning, we have a 2D Gaussian Splatting representation of our scene, that we can use to estimate the motion, in a ``render and compare" strategy.\\
For easy understanding, we will now only consider the center of the Gaussians, and manipulate them only as 3D points. The other Gaussians properties are only useful after the projection of these points; for the rendering.

\subsection{Motion compensation}
\paragraph{LF disparity to 3D.} Without loss of generality, we consider that the observed rigid scene is affected by a constant motion during the acquisition, defined by a 3D rotation $\boldsymbol{\omega}$ and a 3D translation $\mathbf{v}$ in the Euclidean space. This assumption can be modified by choosing diverse motion models (\eg, piecewise constant velocities across the image~\cite{magerand2010generic} or constant acceleration~\cite{dahmouche2008high}).
We can relate the center of the Gaussians $\mathbf{P}$ in the disparity domain to its 3D coordinates in the world coordinates frame similarly to the representation adopted in \cite{hci-dataset, hci-dataset-eval}. The position of a Gaussian center $\mathbf{P}=[X,Y,Z]^t$ in the 3D Euclidean space is computed as:
\begin{equation}\label{lci-eq}
    \begin{split}
	Z & = \beta P_f / (d  P_f  w + \beta),\\
	X & = Z (u-u_0)/f,\\
	Y & = Z (v-v_0)/f,\\
    \end{split}
\end{equation}
where $w$ is the size of the sensor in mm, $P_f$ the distance of the focal point of the LF, $(u_0, v_0)$ the center pixel of the image, $f$ the focal distance $F$ divided by the pixel size, and $b$ the baseline between two images, and $\beta = b F \max(2u_0,2v_0)$.
\paragraph{Link between motion and deformation.} We want to link the deformation created by the 3D rolling shutter effect to the motion. Indeed, given any motion, we can compute the deformation it created on the original (not RS affected) 3D shape, \ie a ``static shape”. The goal is to displace all the Gaussian centers $\mathbf{P}$ at the place they were (or will be) supposed to be at a specific time. In the estimated geometry from the previous stage, each Gaussian was observed at different times. We define the ``central observation time'' $\tau$ as the signed time between the time of projection of $\mathbf{P}$ in the central SAI, and the time of acquisition of the central line of pixel in the SAIs. We define the ``estimated static shape'' $\mathbf{P_s}$ as, given a motion $(\omega, v)$, the position of the point $\mathbf{P}$ at time $\tau = 0$. The relation between $\mathbf{P_s}$ and $\mathbf{P}$ is given by
\begin{equation}
    \mathbf{P}_s(\boldsymbol{\omega}, \mathbf{v}) = 
    \begin{bmatrix}
        \mathbf{R}(-\tau\omega) & -\tau \mathbf{v} \\
        \mathbf{0} & 1
    \end{bmatrix}
    \begin{pmatrix}
        \mathbf{P} \\
        1
    \end{pmatrix},
    \label{eq:deform}
\end{equation}
with $\mathbf{R}(-\tau\omega)$ computed with the Rodrigue's formula.
\paragraph{Reprojection of the estimated static shape.} If the motion $(\boldsymbol{\omega}, \mathbf{v})$ have been correctly estimated, the estimated static shape is correct. To verify this motion hypothesis, we project the points $\mathbf{P}_s$ in the SAIs. We re-use the motion hypothesis to move the ``estimated static shape'' in order to simulated the RS acquisition. We consider that the image formation process is band-wise Global Shutter; we consider a small group of lines that we consider is projected at the same time $\tau_\lambda$. In definitive, we perform our RS projection by assembling multiple parts of multiple GS projection at different times. Then, for a given line group (thus for a given time, $\tau_\lambda$) and given the motion hypothesis $(\boldsymbol{\omega}, \mathbf{v})$, one project the points from their position $\mathbf{P}_\lambda$ that is their position in the ``estimated static shape'' moved with respect to motion and projection time as
\begin{equation}
    \mathbf{P}_\lambda(\boldsymbol{\omega}, \mathbf{v}) = 
    \begin{bmatrix}
        \mathbf{R}(\tau_\lambda\boldsymbol{\omega}) & \tau_\lambda \mathbf{v} \\
        \mathbf{0} & 1
    \end{bmatrix}
    \begin{pmatrix}
        \mathbf{P}_s(\boldsymbol{\omega}, \mathbf{v}) \\
        1
    \end{pmatrix}.
    \label{eq:deform2}
\end{equation}
It is easy to verify that, if we project the points in the central line of SAI, $\mathbf{P}_\lambda(\boldsymbol{\omega}, \mathbf{v}) = \mathbf{P}$, since $\tau_\lambda = \tau$ for any motion.
\paragraph{Error estimation and optimization.} To summarize the previous steps, we are now able to render an image of the 3D RS corrected shape re-imaged with an RS acquisition. We recall that a given Gaussian is not imaged at the same time in SAI of different rows. During the second stage, we optimize the 6D motion parameters by minimizing the error of the SAI rendered using the Gaussians with the measured SAIs. Given a 4D Light Field $\mathbf{LF}$, a viewpoint $(x,y)$, and a group of lines $\mathbf{k_\lambda}$ spanning from line $v$ to the line $v+\lambda$, we can render an image using the Gaussians with centers $\mathbf{P}_\lambda$. We note this rendered image $\mathbf{LF_R}(x,y,\mathbf{k_\lambda}, \boldsymbol{\omega}, \mathbf{v})$. Then the error of reprojection $e$ we minimize is given by: 
\begin{equation}
    e(\boldsymbol{\omega}, \mathbf{v}) = \lVert\mathbf{LF_R}(x,y,\mathbf{k_\lambda},\boldsymbol{\omega}, \mathbf{v})-\mathbf{LF}(x,y,\mathbf{k_\lambda})\rVert. \label{eq:loss}
\end{equation}
During the optimization, we go through different views $(x,y)$ and different line groups $\mathbf{k_\lambda}$ and we minimize the error in \cref{eq:loss}, only optimizing for the motion parameters. After optimization, we perform rendering using the learned Gaussians in a global shutter manner in order to obtain the estimated RS compensated central view and disparity map.
\vspace{-0.5cm}

\section{RSLF+ dataset}
\begin{figure}[t]
    \centering
    \includegraphics[width=0.75\linewidth]{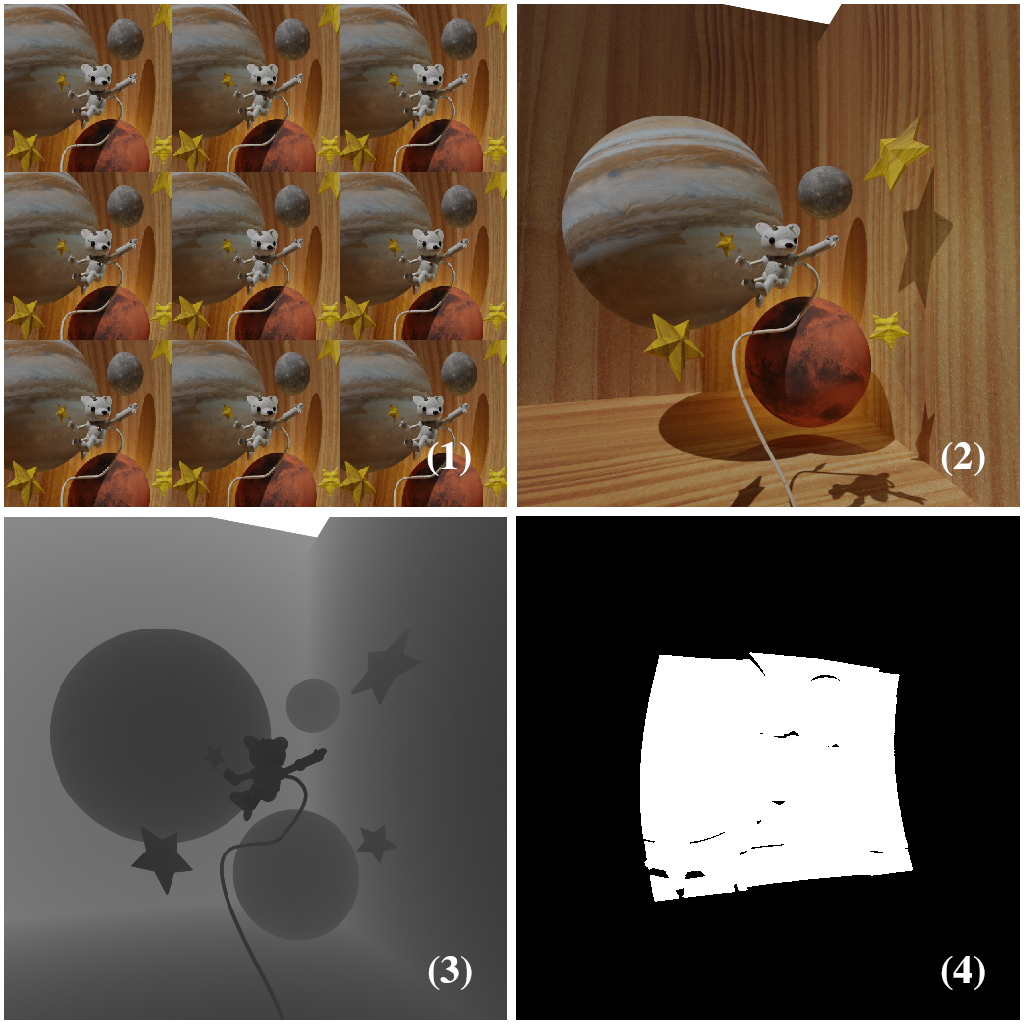}
    \caption{\textbf{One sample from the RSLF+ dataset}. For each scene and motion, we provide: \textbf{(1)} One LF of dimension $9\times9\times512\times512$ with rolling shutter effect (here is shown only a $3\times3$ sample) \textbf{(2)} The central SAI with the double field of view \textbf{(3)} The depth map with the double field of view \textbf{(4)} The mask of visible pixels adapted to the motion.}
    \label{fig:RSLF+_dataset}
\end{figure}

\begin{figure}[t]
    \centering
    \includegraphics[width=0.95\linewidth]{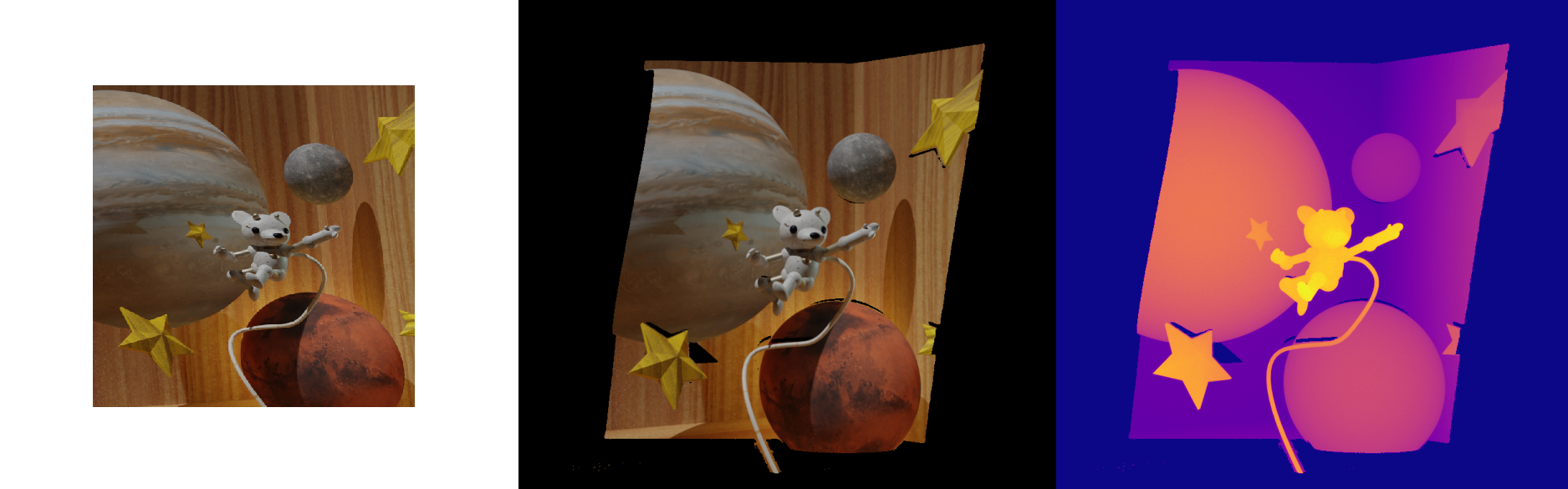}
    \caption{\textbf{SAI deformed with the RS effect}. The ground truth image (center) and disparity (right) uses the visibility masks to consider only the visible points in the RS affected central SAI (left).}
    \label{fig:masks}
\end{figure}

The work~\cite{HermesWACV24} proposed the first RSLF dataset in order to allow the evaluation of the rolling shutter effect with LFs. We propose here the RSLF+ dataset, a highly inspired dataset of this seminal work. The proposal of this dataset was motivated to allow evaluation with more textured scenes and by explicitly handling pixel occlusions (occlusion masks). \cref{fig:RSLF+_dataset} shows a sample of the proposed synthetic dataset. The scenes are textured and with different level of geometric details: from a flat object as checkerboards, to highly detailed objects such as vegetation (please see the bottom row in \cref{fig:results}). The motion scenarios are different for every scenes, and are spread seemingly regularly throughout the whole six dimensions of velocity space. We provide not only the depth map with twice the field of view, but also the central view with twice the field of view, so that we can compare the estimated reconstructions and the ground truth even with fast motions.\\
More importantly, we also provide visibility masks that indicate which pixels in the ground truth are seen in the RS deformed scenes. 
This allows us to evaluate more rigorously the reconstructions, since we can consider the points outside of the field of view of the static central SAIs, and also we do not consider pixels that are outside the field of view of the moving central SAIs. \cref{fig:masks} shows the ground truth adapted to the velocities, thanks to these visibility masks.
The dataset comes with an automatic evaluation tool providing the metrics shown in \cref{tab:errors,tab:errors+}. 

\begin{table*}[t!]
	\centering
	\resizebox{0.7\linewidth}{!}{%
   \begin{tabular}{lllllllllllll}
& \multicolumn{3}{c}{abs diff $\downarrow$} & \multicolumn{3}{c}{rmse $\downarrow$} & \multicolumn{3}{c}{$\delta < 1.25 \uparrow$}\\
\cmidrule(lr){2-4} \cmidrule(lr){5-7} \cmidrule(lr){8-10}
 & GS & slow & fast & GS & slow & fast & GS & slow & fast\\
\cmidrule(lr){2-4} \cmidrule(lr){5-7} \cmidrule(lr){8-10}

Jeon-CVPR \cite{jeon2015accurate} & 0.373 & 0.384 & 0.499 & 0.493 & 0.521 & 0.674 & 0.47 & 0.485 & 0.378\\
OACC-Net \cite{OACC-Net} & 0.688 & 0.702 & 0.826 & 0.824 & 0.831 & 0.945 & 0.524 & 0.514 & 0.425\\
RSLF-d \cite{HermesWACV24} & 0.342 & 0.361 & 0.413 & 0.459 & 0.506 & 0.573 & 0.43 & 0.451 & 0.421\\
Ours & \textbf{0.309} & \textbf{0.32} & \textbf{0.363} & \textbf{0.43} & \textbf{0.442} & \textbf{0.516} & \textbf{0.604} & \textbf{0.588} & \textbf{0.509}\\
\bottomrule
\end{tabular}
  }
 \caption{Quantitative results on the RSLF dataset~\cite{HermesWACV24}. Mean over the scenes, separated in the 3 motion categories.}
 
	\label{tab:errors}
\end{table*}

\begin{table*}[t!]
	\centering
	\resizebox{0.7\linewidth}{!}{%
   \begin{tabular}{lllllllllllll}
& \multicolumn{3}{c}{abs diff $\downarrow$} & \multicolumn{3}{c}{rmse $\downarrow$} & \multicolumn{3}{c}{$\delta < 1.25 \uparrow$}\\
\cmidrule(lr){2-4} \cmidrule(lr){5-7} \cmidrule(lr){8-10}
 & GS & slow & fast & GS & slow & fast & GS & slow & fast\\
\cmidrule(lr){2-4} \cmidrule(lr){5-7} \cmidrule(lr){8-10}

Jeon-CVPR \cite{jeon2015accurate} & 0.064 & 0.094 & 0.147 & 0.148 & 0.249 & 0.362 & 0.874 & 0.857 & 0.808\\
OACC-Net \cite{OACC-Net} & 0.052 & 0.084 & 0.129 & 0.176 & 0.263 & 0.338 & 0.966 & 0.929 & 0.867\\
RSLF-d \cite{HermesWACV24} & 0.114 & 0.108 & 0.111 & 0.281 & 0.26 & 0.274 & 0.793 & 0.799 & 0.806\\
Ours & \textbf{0.03} & \textbf{0.041} & \textbf{0.06} & \textbf{0.133} & \textbf{0.161} & \textbf{0.208} & \textbf{0.97} & \textbf{0.958} & \textbf{0.93}\\
\bottomrule
\end{tabular}
  }
\caption{Quantitative results on the RSLF+ dataset. Mean over the scenes, separated in the 3 motion categories.\vspace{-0.5cm}}
 
\label{tab:errors+}
\end{table*}
\section{Experiments}
\paragraph{Setup and optimization.} All the components of the method are implemented in Pytorch and the optimization is done with Adam. In the first step, we optimize five parameters per Gaussian, and we use 20 000 Gaussians totaling 100 000 parameters. Since we render the scene band after band (group of lines), we do not optimize all the parameters at once. In the second step, we only optimize the velocity parameters. The computation time is about 10 minutes per scene for the whole pipeline with a 16GB graphic card. We believe this time can be strongly reduced, since only 100 iterations are needed at each step, and as an important part of the computation time is spent looping between rendering tiles that can be parallelizable. In order to lighten the computation (and to ease the optimization process), we use 5 (over 9) images of the central line of SAI in the first step, and only 4 (over 81-9 = 72) images in the second step. These images correspond to the extreme ``corners" of the array of SAI.
\paragraph{Metrics and datasets.} The RSLF dataset~\cite{HermesWACV24} is, up to our knowledge, the only available dataset that contains light fields with RS effect. It contains different scenes with eleven different motions (including a static configuration, five ``slow" motions, and five ``fast" motions). As the study of rolling shutter light fields is at its premises there is no existing public dataset containing LFs acquired with a real sensor with rolling shutter effect. Yet, since our method also has the ability to work on LF images with no rolling shutter effect, we also perform evaluations on existing datasets with global shutter LFs. Therefore we tested our method on the 4D LF dataset~\cite{honauer2016benchmark}, which is a synthetic dataset that allows for quantitative evaluation as it provides ground truth depth. We also tested our method on LFs acquired with real plenoptic cameras, from three different datasets used for rendering (\cite{rerabek2016new}, \cite{Inria_LF_dataset}, \cite{pertuz2018focus}), however, they do not provide ground truth depth information and we can only provide qualitative assessment of the learned geometry. 
The quantitative results are then computed on the 4D LF dataset~\cite{honauer2016benchmark}, the RSLF dataset~\cite{HermesWACV24} and our proposed dataset (RSLF+ dataset). We have used three classical metrics for the evaluation: abs diff (the mean of the absolute differences between the images and ground truth), rmse (the root mean squared error) and $\delta < 1.25$ (the proportion of the errors under 25\% of the ground truth, i.e. the proportion of error under 0.25 in relative difference). For the evaluation on the 4D LF dataset, we used the evaluation toolbox provided by the authors. The metrics chosen for this evaluation are MSE (mean squared error) and bad pix (proportion of pixel under a threshold in absolute error).

\noindent\textbf{Baselines.} The method was compared with three recent algorithms from the literature that deal with depth estimation from LF: a handcrafted depth estimation method~\cite{jeon2015accurate} and a learning-based method~\cite{OACC-Net}. To complement these experiments, we also compare to a densified version our closest competitor~\cite{HermesWACV24}, that is, up to our knowledge, the only method to deal with RS effect in LFs. This densification was done by projecting the resulting sparse point cloud and performing linear interpolation to the closest projected points, inside the convex hull of the projected points. We named this method RSLF-d. In \cref{fig:hci}, we also compare our method with EPI1 \cite{johannsen2016sparse}, EPINET \cite{shin2018epinet} and LF-OCC \cite{wang2015occlusion}.

\begin{figure*}[t]
    \centering
    \includegraphics[width=0.8\linewidth]{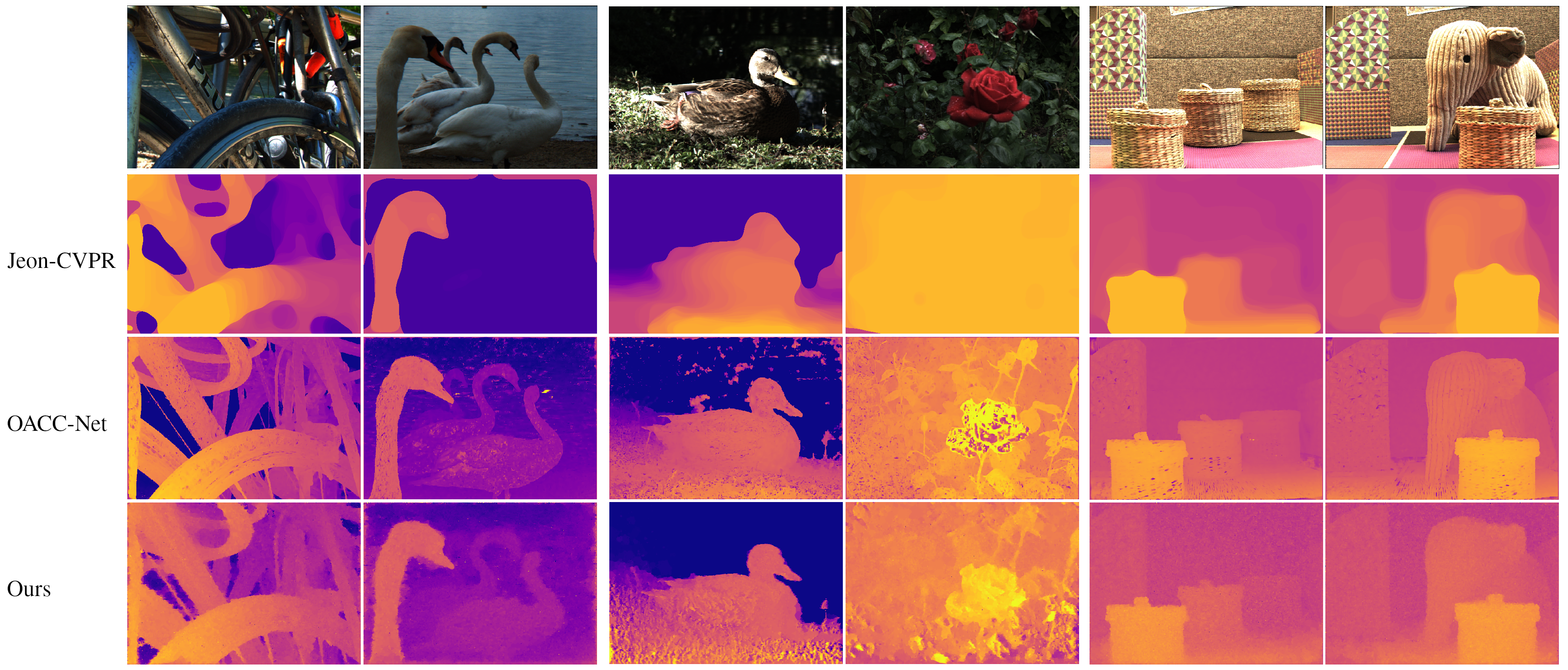}
    \caption{Qualitative evaluation on real LF data. The upper row shows the real central SAI. The first two columns are images from the EPFL Light-field data set \cite{rerabek2016new}, the two central columns are images from the Inria dataset \cite{Inria_LF_dataset} and the last two columns are images from the CVIA dataset \cite{pertuz2018focus}. The disparity maps are estimated by, in order, Jeon-CVPR~\cite{jeon2015accurate}, OACC-Net~\cite{OACC-Net} and Ours.}
    \label{fig:real_data}
\end{figure*}

\begin{figure}[t]
    \centering
    \includegraphics[width=\linewidth]{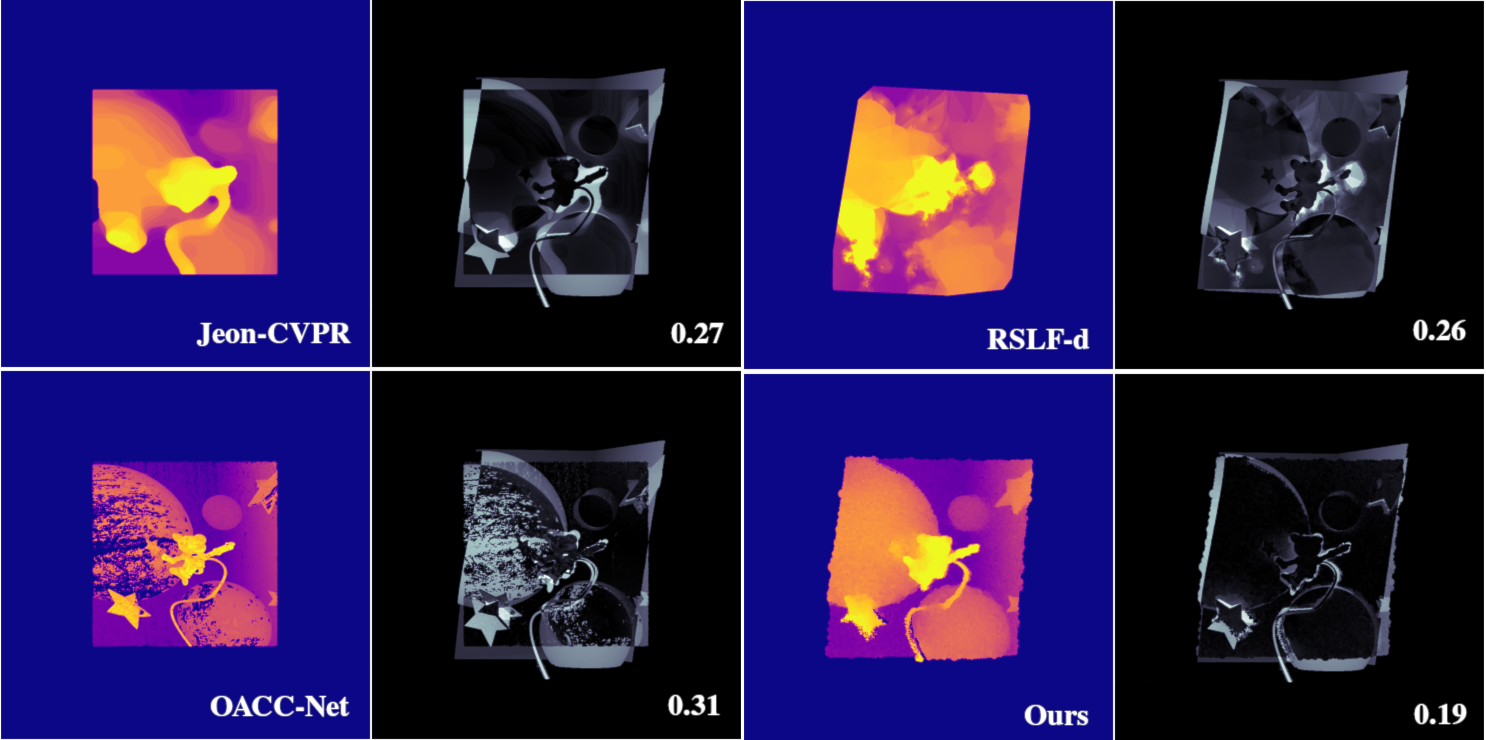}
    \caption{Comparisons on a scene with RS effect for the 4 methods considered. On the left, the computed disparity maps, and on the right, the RMSE depth errors. Brighter intensity indicates a higher error.}
    \label{fig:compare}
\end{figure}

\begin{figure}[t]
    \centering
    \includegraphics[width=\linewidth]{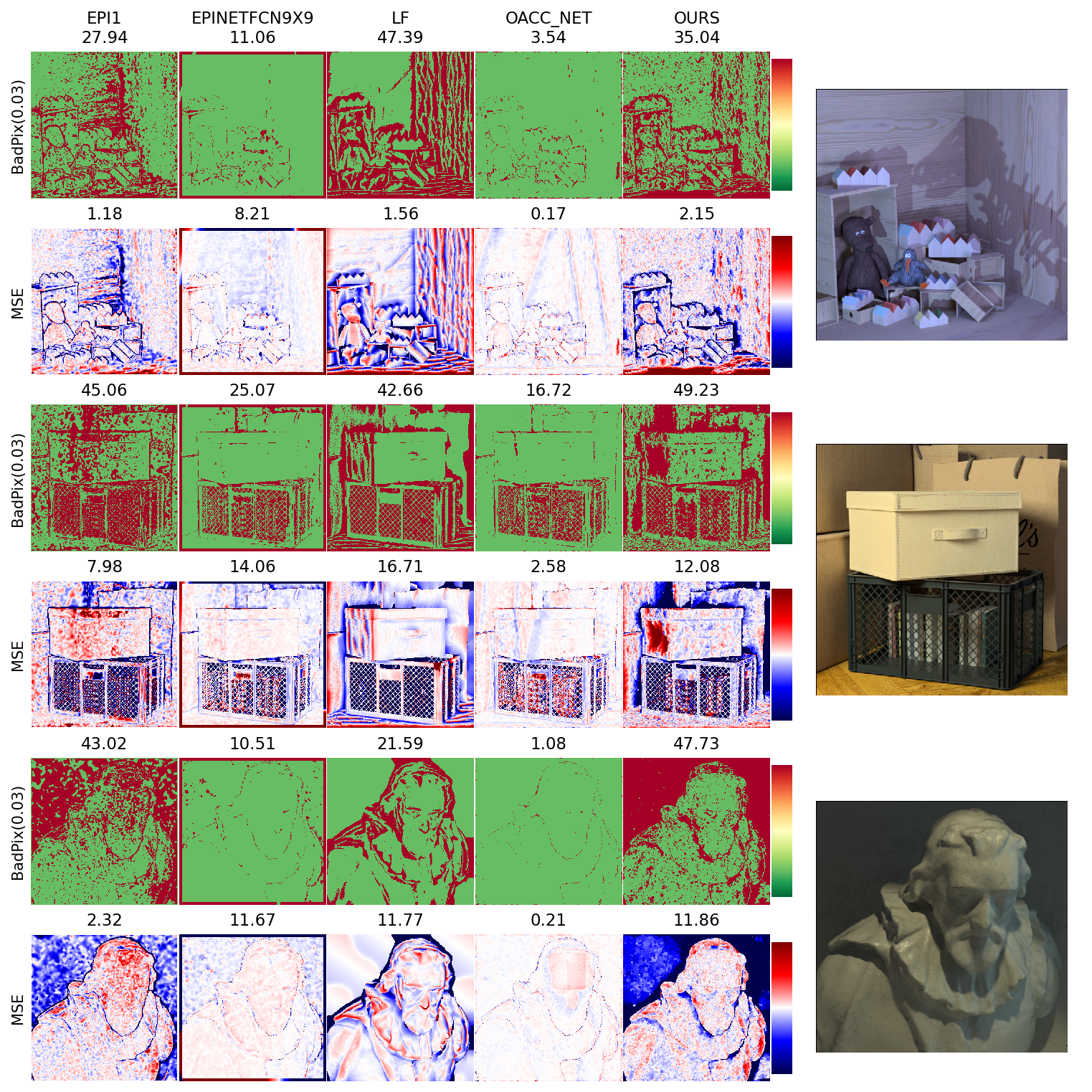}
    \caption{Comparisons with ``static" scenes (if we consider a RS acquisition) with comparable performance. The main errors appear in low textured regions. Note for example the solid background in the third scene which could be hardly computed without priors.}
    \label{fig:hci}
\end{figure}
\begin{figure*}[t]
    \centering
    \includegraphics[width=0.7\linewidth]{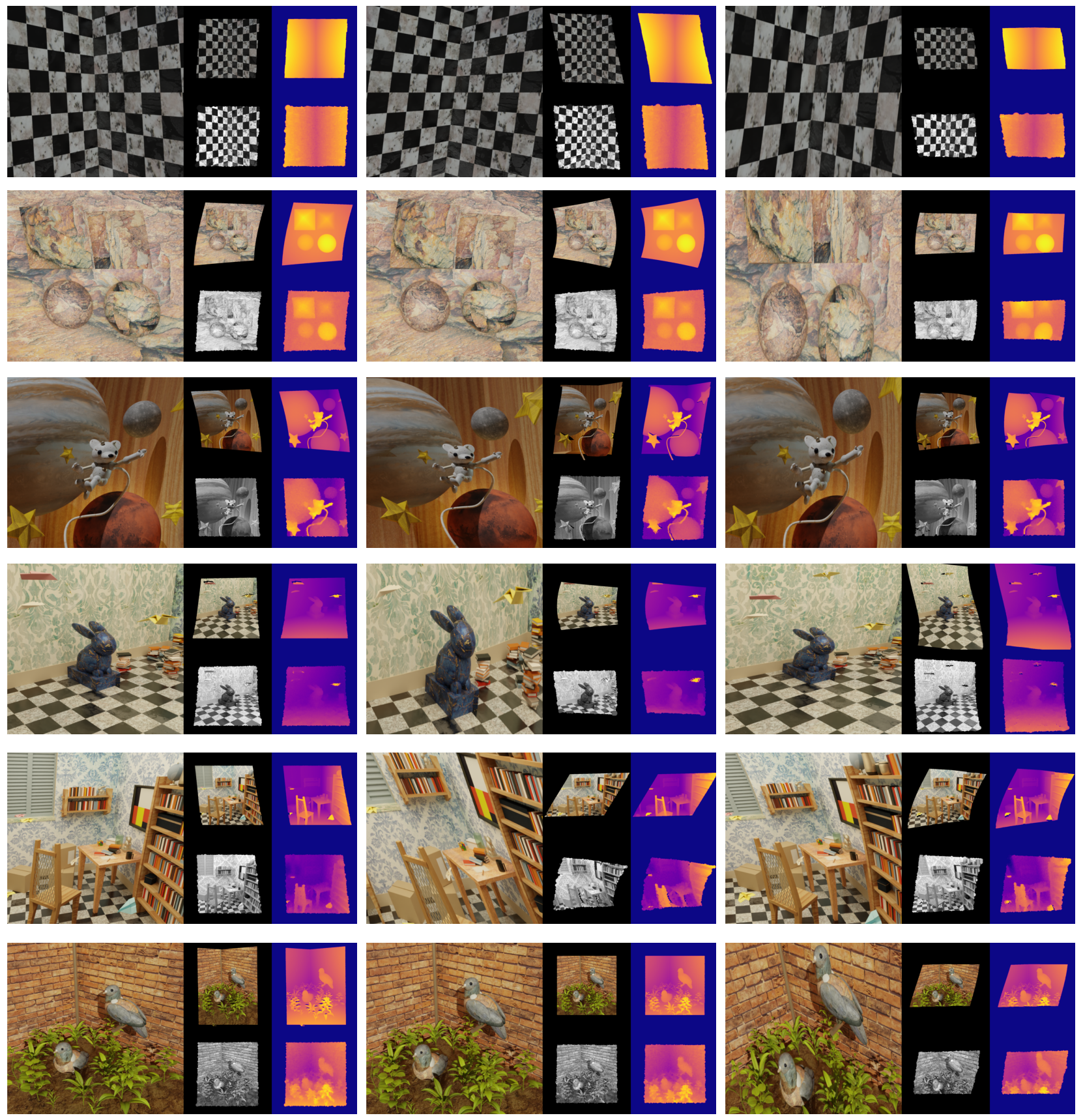}\vspace*{-0.3cm}
    \caption{Qualitative results on the RSLF+ dataset. \textbf{In big}: the central SAIs. For each scene, we show 3 of the 6 motion scenarios. \\
    \textbf{In small}: \textit{above}: The ground truth appearance and disparity map, \textit{under}: The estimated appearance and disparity map.}
    \label{fig:results}\vspace*{-0.1cm}
\end{figure*}
\noindent\textbf{Results.}
The quantitative results on the RSLF dataset~\cite{HermesWACV24} are shown in \cref{tab:errors}. Note that this comparison is done in our disfavor, since we only compute the differences inside the field of view of the central SAI, although our method is able to compute intensity and disparity outside of this field of view as discussed in \cref{intro}. We observe that the proposed method outperforms the competitors even in the static case (named GS, for Global Shutter) in \cref{tab:errors}. These results are explained by the fact that the disparity maps produced by the handcrafted depth estimation method~\cite{jeon2015accurate} are ``smooth" (see \cref{fig:compare}), and that the learned depth estimation method seems to struggle to generalize. OACC-net present some artefacts (see \cref{fig:compare}), although it produced state-of-the-art disparity maps in the 4D LF dataset~\cite{honauer2016benchmark}, as shown in \cref{fig:hci}. More importantly, please notice how RSLF-d and our method are less affected by the RS, and how our method performs systematically better than RSLF-d, thanks to the depth density of our method, that better handles fine geometrical details. The quantitative results on the RSLF+ dataset are shown in \cref{tab:errors+}. Note that the results show improved performances for all methods considering the depth metrics, notably due to the more textured scenes, while our method still outperforms all the baselines. Some qualitative results are shown in \cref{fig:results}. Note how the estimated shapes bends in the right manner compared to the ground truth, in order to correct the RS effect, visible in the central SAIs. We also include some qualitative results of the methods on real LFs images in \cref{fig:real_data}. We see that we perform comparably to the methods from the literature in the context of global shutter scenes. While not being as sharp as the state-of-the art technique \cite{OACC-Net}, we show comparable results for the modern standards of depth from LF, while not being designed for a global shutter. Note that we are only using the central row of SAI to compute this disparity, as explained in \cref{sec:c_row}. Some qualitative results for the different methods evaluated in the the proposed dataset are shown in \cref{fig:compare}. Please notice how our method compensate the deformation and how the generated depth map is more precise than RSLF-d. Visualization of the results on the 4D LF dataset are shown in \cref{fig:hci}. These results suggest the proposed method perform comparably to the other methods of the benchmark, struggling mostly in low-textured regions.

\paragraph{Ablation Study.}
We perform an ablation study of our method, in order to evaluate the pertinence of each step, namely, the fine-tuning of the Gaussian Splatting representation of the RS agnostic initialization (named ``init" in the ablation study), and the motion estimation (named ``motion" in the ablation study). Results are shown in \cref{tab:ablation} and show how the complete method performs better in the scenarios with rolling shutter effect. We also compare in terms of appearance, with the metric ``rmse (intensity)". This metric is not shown in the main tables of the paper because the other methods   shown it in the previous tables do not provide an estimate of the appearance of the scene. The small reduction of quality in the static (GS) scenarios  compared with the method with no motion estimation, displayed in \cref{tab:ablation}, is explained by the fact that the full method is under-constrained, and that it will produce more error by trying to refine the initial scene estimation, with a wrong (small) motion estimation.

\begin{table}[t!]
	\centering
	\resizebox{\linewidth}{!}{%
\begin{tabular}{lllllll}
& \multicolumn{3}{c}{rmse (intensity) $\downarrow$} & \multicolumn{3}{c}{abs diff (depth) $\downarrow$}\\
\cmidrule(lr){2-4} \cmidrule(lr){5-7}
& GS & slow & fast & GS & slow & fast\\
\cmidrule(lr){2-4} \cmidrule(lr){5-7}
No init, no motion & 0.061          & 0.109          & 0.154          & 0.032          & 0.057          & 0.101         \\
No motion          & \textbf{0.059} & 0.112          & 0.155          & \textbf{0.029} & 0.054          & 0.102         \\
No init            & 0.105          & 0.112          & 0.12           & 0.045          & 0.053          & 0.069         \\
Full               & 0.063          & \textbf{0.088} & \textbf{0.117} & 0.03           & \textbf{0.041} & \textbf{0.06}\\
\bottomrule
\end{tabular}
}
\caption{Quantitative results of the ablation study on the RSLF+ Dataset. Mean over the scenes, separated in the 3 motion categories.} 
\label{tab:ablation}\vspace*{-0.3cm}
\end{table}

\section{Conclusion}
We present in this paper a rolling shutter aware dense scene reconstruction method adapted to light-field images. We proposed a representation of the scene based on 2D Gaussians, that leverages the specificity of a light field captured with a plenoptic camera, allowing to obtain a dense reconstruction without requiring correspondences. We designed this representation to properly model the relation between rolling shutter deformation and camera motion in this context, and we show the capability of the proposed method to compensate for rolling shutter effects during the scene geometry and appearance estimation. We believe that this rich information extraction and modelling from a single capture have, more than its abstract interest, strong potential in vision tasks. The source code and dataset will be made publicly available.

\paragraph{Acknowledgements.} The authors would like to thank the funding from the French ``Investissements d’Avenir'' program, project ISITE-BFC, contract ANR-15-IDEX-03, by the Conseil Régional BFC from the project ANER-MOVIS, by ANR
(ANR-23-CE23-0003-01) and by ``Grand Prix Scientifique 2018, Fond. Ch. Defforey-Institut de France".

\cleardoublepage

{\small
\bibliographystyle{ieee_fullname}
\bibliography{egbib}
}

\cleardoublepage

\twocolumn[  
\begin{@twocolumnfalse}
    \begin{center}
        \section*{[Supplementary Material - WACV 2025]:\\ Dense Scene Reconstruction from Light-Field Images Affected by Rolling Shutter}
        \vspace{1cm}
    \end{center}
\end{@twocolumnfalse}
]  

\setcounter{section}{0}
\setcounter{figure}{0}

In this ``Supplementary Material'' to our paper, we provide an introduction to light field theory and light field (plenoptic) cameras. We also discuss the natural synchronization on the SAIs and the creation of RSLF-d from the adaptation of \CR{our} seminal work [2]. The objective of this ``Supplementary Material'' is notably to provide key ideas for a better understanding of the unique properties of the Rolling-Shutter Light Fields.
\vspace{1cm}
\section{The plenoptic camera and the light field: an overview} \label{sec:overview}
To facilitate the understanding of the paper, we provide in this section an brief overview of concepts on plenoptic cameras and light field theory.

\paragraph{Plenoptic camera optical system.} 
\begin{figure}[h]
    \centering
    \includegraphics[width=0.9\linewidth]{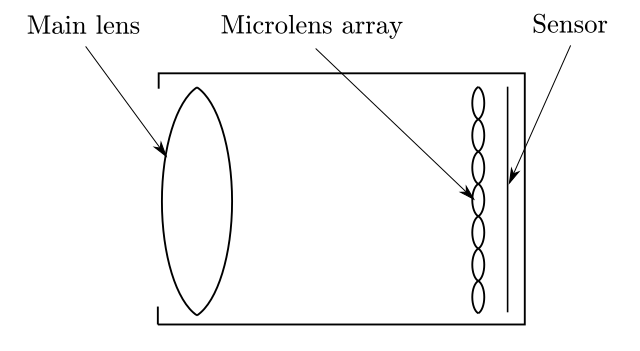}
    \caption{A simplified plenoptic camera scheme. The main difference to a standard camera is the addition of a microlens array between the main lens and the sensor.}
    \label{fig:pleno_schema}
\end{figure}
In 1992, Adelson and Wang introduced the so called ``plenoptic camera'', an optical design capable of capturing a 4D light field in a single acquisition. It was inspired by 1908's integral photography of Lippmann [1]. In the following years, this optical sensor gained popularity in the computer vision community, sometimes under the term ``light field cameras'' (popularized by the monopolistic company that produce these cameras today). Plenoptic cameras capture the light in 4D spatial dimensions (with an angular resolution superior to one, contrary to monocular cameras). In order to do that, it uses an optical system composed of a classical part (``the main lens'') and a micro-lens array in front of the sensor (between the main lens and the sensor). The model of the plenoptic camera is shown in \cref{fig:pleno_schema}. There are multiple interpretation of the way this imaging system creates images. The easier to understand it, according to us, is to see every single microlens as the end of a unique optical system that images the scene like a tiny camera. The whole array of microlens can then be seen as an array of cameras that have a resolution of a dozen of pixel. Another interpretation, is the idea that the microimages (the images behind the microlenses) capture the angular incidence of the light rays at the position of the microlens. That is why one can refer to the supplementary dimensions (in comparison to monocular cameras) as the ``angular resolution''. The images obtained with such a camera look like the one presented in \cref{fig:pleno_pict}.

\begin{figure}[t]
    \centering
    \includegraphics[width=\linewidth]{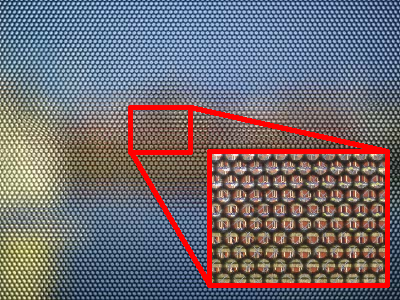}
    \caption{The unprocessed image obtained with a plenoptic camera. We can see in the detail the multitude of micro-images produced by the microlenses. Images from [3].}
    \label{fig:pleno_pict}
\end{figure}

\begin{figure*}[t]
    \centering
    \includegraphics[width=0.95\linewidth]{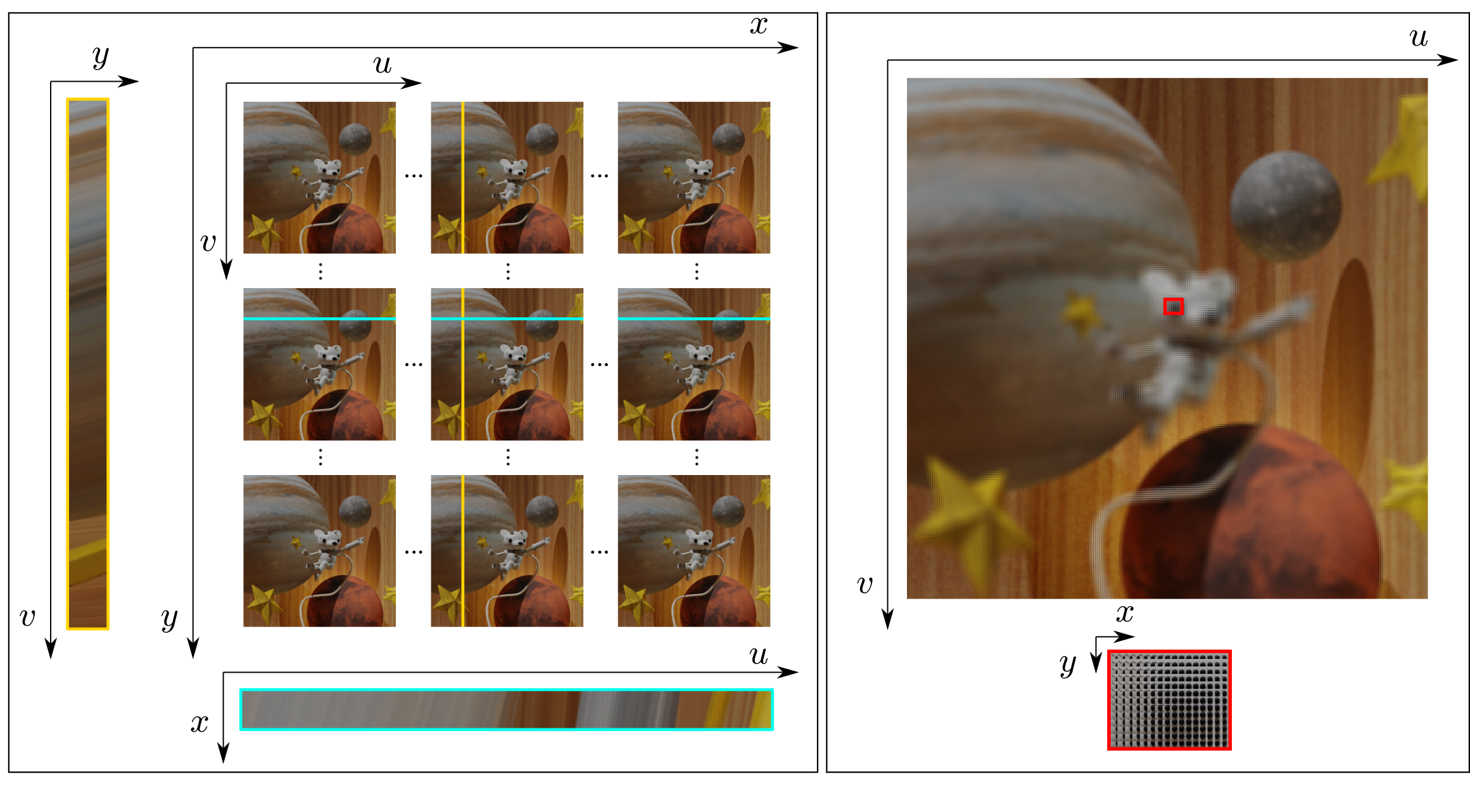}
    \caption{The representations of the 4D light field. In the paper, we use the array (x,y) of SAI (u,v) as an equivalence of an array of cameras.}
    \label{fig:4D_data}
\end{figure*}
\paragraph{Light Field representation.} In order to exploit such an image we first have to decode the light field out of it. In this context, decoding means segmenting the small circular images called the micro-images and constructing a 2D array of 2D micro-images, thus a 4D tensor, that is typically referred as 4D light field in the context of plenoptic cameras (the light field is in fact a broader concept in radiometry). We name the dimensions of the 4D LF with $(x,y,u,v)$. $(x,y)$ give the coordinates of the ``viewpoint''. They are related to the coordinates inside the micro-images. These are the aforementioned ``angular resolution''. (u,v) give the coordinates in pixels inside an image from a given viewpoint, named the ``sub-aperture image'' (SAI). (u,v) are the ``spatial resolution''. These SAI are equivalent to images acquired with an array of cameras with very small baselines. One can also create images by cutting the 4D data in the (x,u) plane or (y,v) plane, called the epipolar plane images. \cref{fig:4D_data} shows the way the 4D light field can be represented.

\section{Time synchronization of the SAIs}
We have shown in \cref{sec:overview} of this ``Supplementary Material'' how the LF is extracted from the plenoptic image. In order to introduce RS we just need to consider that the lines of pixels are acquired sequentially. As the micro-images are very small in comparison to the whole picture, we make the reasonable assumption that all the pixels of a micro-images are acquired at the same time. We can then assume that the lines of micro-images are acquired sequentially, but that they are locally global shutter. What this fair modelization means is that the relation between the time of acquisition of a pixel at coordinates (x,y,u,v) (See Fig. 3) is only related to its v coordinate. As the Sub-Aperture Images are expressed in the same (u,v) coordinate system, they begin to be acquired at the same time (same v) and finish to be acquired at the same time. The SAIs are then naturally synchronized. (Rigorously, the most medium time one can observe between two SAI is the time between the acquisition of the first line of pixel in a micro-image, and the acquisition of the last line of pixel in the same micro-image. A few lines. And thus a short time; that we consider negligible in our study and that, in further study would be, at worst, measurable.)

\section{The motion information in RSLF}
\begin{figure}[t]
\centering
\includegraphics[width=0.7\linewidth]{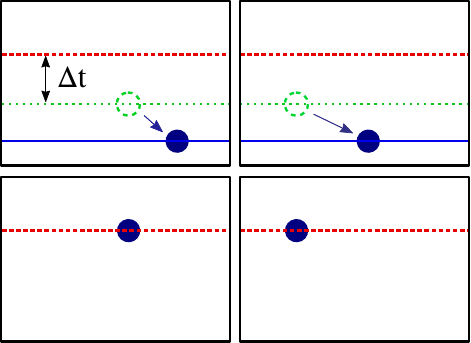}
\caption{The motion estimation is visible in 3D for every point, since we can observe them from different view points (change in SAI) and at different times (change in vertical coordinate inside the SAI).}
\label{fig:key}
\end{figure}
In RS LF, there is no prior needed to compute the instantaneous velocity of each point in 3D. Indeed, we observe every points at different times and from different angles. It is this chore property of the RS LF that we exploit. \cref{fig:key} shows 4 adjacent SAIs. The blue points are the projections of the same 3D point in these SAI. The red dotted line is the line acquired at the time at which the point is projected in the bottom row of SAI. Because of parallax, the point is not projected in the same line in the upper row. In the GS case, it would be projected in the position of the green dotted circles. But since we are in RS, a time $\delta t$ has passed between the red dotted line and the green dotted line. The 3D point moved in space during this time. It is then projected in another line (full blue line).

\begin{figure}[t]
    \centering
    \includegraphics[width=0.8\linewidth]{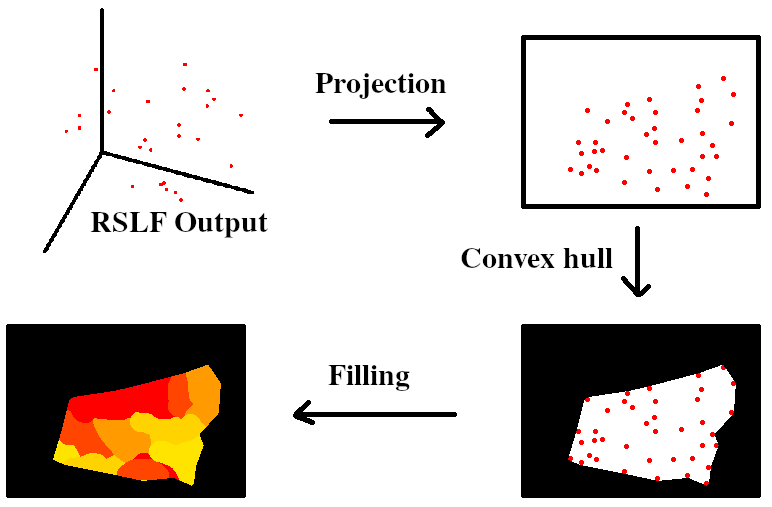}
    \caption{Overview of the densification of the original RSLF result.}
    \label{fig:RSLF-d}
\end{figure}

\section{RSLF-d: the densified version of RSLF}
In order to ease the comparison between the different methods, we propose a new take on~[2] by densifying it. Indeed, this method returns a point cloud that is hard to compare with ground truth. It would also need its own metrics, that would add complexity to the comparison. That is why we propose RSLF-d (that can be seen as an annex, minor contribution). By projecting the point cloud obtained with their method (we set it up to have a lot of points) we obtain a very sparse depth map. We fill out the empty areas with the closer value. Also, we compute the convex hull of the 2D projected point cloud in order to restrict the estimated depth map inside the hull of estimated points (we recall that we compare with masked ground truth). The pipeline from RSLF-d is presented in \cref{fig:RSLF-d}. By doing this, we make sure to stay relatively true to the work of [2], while allowing us to use the evaluation tools we crafted.

\small{
\section*{References}
\begin{itemize}
    \item[]\!\!\!\!\!\!\!\!\!\!\![1]~Gabriel Lippmann. La photographie integrale. \textit{Comptes-Rendus}, 146:446–451, 1908.
    \item[]\!\!\!\!\!\!\!\!\!\!\![2]~Hermes McGriff, Renato Martins, Nicolas Andreff, and Cedric Demonceaux. Joint 3D shape and motion estimation from rolling shutter light-field images. In \textit{WACV}, 2024.
    \item[]\!\!\!\!\!\!\!\!\!\!\![3]~~Mats Wernersson. The camera maker.
\end{itemize}

\end{document}